# The Lumière Project: Bayesian User Modeling for Inferring the Goals and Needs of Software Users

Eric Horvitz, Jack Breese, David Heckerman, David Hovel, Koos Rommelse[†]
Microsoft Research
Redmond, WA 98052-6399
{horvitz,breese,heckerma,davidhov}@microsoft.com

## Abstract

The Lumière Project centers on harnessing probability and utility to provide assistance to computer software users. We review work on Bayesian user models that can be employed to infer a user's needs by considering a user's background, actions, and queries. Several problems were tackled in Lumière research, including (1) the construction of Bayesian models for reasoning about the time-varying goals of computer users from their observed actions and queries, (2) gaining access to a stream of events from software applications, (3) developing a language for transforming system events into observational variables represented in Bayesian user models, (4) developing persistent profiles to capture changes in a user's expertise, and (5) the development of an overall architecture for an intelligent user interface. Lumière prototypes served as the basis for the *Office Assistant* in the Microsoft Office '97 suite of productivity applications.

## 1 Introduction

Uncertainty is ubiquitous in attempts to recognize an agent's goals from observations of behavior. The Lumière project at Microsoft Research has been focused on leveraging methods for reasoning under uncertainty about the goals of software users. At the heart of Lumière research and prototypes are Bayesian user models that capture the uncertain relationships among the goals and needs of a user and observations about program state, sequences of actions over time, and words in a user's query. Motivating problems include the computation and use of probability distributions over a user's goals for providing appropriate assistance or automated services, for dynamically tailoring language models in speech recognition, and for appropriately guiding the allocation of computational resources in an operating system.

We present challenges with the construction of key components of the Lumière/Excel prototype for the Excel spreadsheet application. The Lumière project was initiated in 1993, and the initial Lumière/Excel prototype was first demonstrated to the Microsoft Office division in January 1994. Research on new applications and extensions continued in parallel with efforts to integrate portions of the prototype into commercially available software applications. In January 1997, a derivative of Lumière research shipped as the *Office Assistant* in Microsoft Office '97 applications.

We will describe key issues with user modeling and provide an overview of Bayesian reasoning and decision making about user needs. Then, we will discuss user modeling for identifying the needs of users working with desktop productivity software. We will review the components and overall architecture of the Lumière system. Finally, we will discuss the influence of Lumière research on software products.

## 2 Bayesian User Models

We shall focus on the use of Bayesian networks and influence diagrams in embedded applications to make inferences about the goals of users–and to take ideal actions based on probability distributions over these goals. We found that Bayesian models can be effective in diagnosing a user's needs and can provide useful enhancements to legacy software applications when embedded within these programs. Additionally, Bayesian user models can provide an infrastructure for building new kinds of services and applications in software.

To date, graphical probabilistic models have been employed largely for such diagnostic tasks as computing the likelihood of alternate diseases in patients or disorders in machines (Horvitz, Breese & Henrion, 1988; Heckerman, Horvitz & Nathwani, 1992; Heckerman, Breese & Rommelse, 1995). However, there has been growing interest in the application of Bayesian and decision-theoretic methods to the task of modeling the beliefs, intentions, goals, and needs of users (Jameson, 1996; Horvitz, 1997). Such *user modeling* problems typically are dominated by uncertainty.

Representations of probability and utility have been explored previously in a variety of user-modeling applications. Models of user expertise and abilities have been used in the context of custom-tailoring the behavior of Bayesian decision-support systems to users. For example, multiattribute utility models were employed in early versions of the Pathfinder pathology diagnostic system for pathology to custom-tailor question asking and explanation to users with differing levels of expertise (Horvitz, Heckerman, Nathwani & Fagan, 1984). Bayesian networks were employed as user models in the Vista system to model the interpretation of patterns of evidence by flight engineers at the NASA Mission Control Center. In that application, concurrent inference with the user and expert models is used to select the most valuable information to display (Horvitz & Barry, 1995). In the realm of modeling the goals of users, temporal probabilistic models of a pilot's goals were explored by Cooper, et al., guided by the challenge of custom-tailoring information displayed to pilots of commercial aircraft (Cooper, Horvitz, Heckerman & Curry, 1988). Probabilistic models have been explored as a representation for recognizing commonsense plans (Charniak & Goldman, 1993), for making inferences about the goals of car drivers in navigating in traffic (Pynadath & Wellman, 1995), and for predicting actions in a multiuser computer game (Albrecht, Zukerman, Nicholson & Bud, 1997). Interest has also been growing steadily on applications of Bayesian user modeling in educational systems. In particular, there have been efforts to model competency and to diagnose problems with understanding computer-based tutoring (Conati, Gertner, VanLehn & Druzdzel, 1997).

---

[†]Presently at Applicare Medical Imaging B.V., Box 936, 3700 AX Zeist, Netherlands.

The high-level goals of the Lumière project are captured by the influence diagram displayed in Figure 1 which represents key aspects of user modeling and automated assistance in computing applications. The decision model considers a user's goals and needs. *Goals* are target tasks or subtasks at the focus of a user's attention. *Needs* are information or automated actions that will reduce the time or effort required to achieve goals. In some situations, we can exploit a predetermined mapping between goals and needs. However, we typically must consider uncertainties and dependencies among these and related variables. As indicated by the variables and dependencies in the influence diagram, a user's acute needs are influenced by the acute goals as well as the user's competency with using software. Prior assistance in the form of online help and the user's background representing experience with computing applications influence the user's competence. A user's needs directly influence or *cause* patterns of activity that might be sensed by watching a user's activity. Such activity includes sequences of user actions recorded as the user interacts with a mouse and keyboard as well as visual and acoustic clues that might be sensed by a video camera and microphone. A user's goals also influence the set of active documents and the presence, instantiation, and display of data structures (*e.g.,* Does a user-authored chart exist in a spreadsheet project? Is the chart currently visible? Does it currently have system focus?).

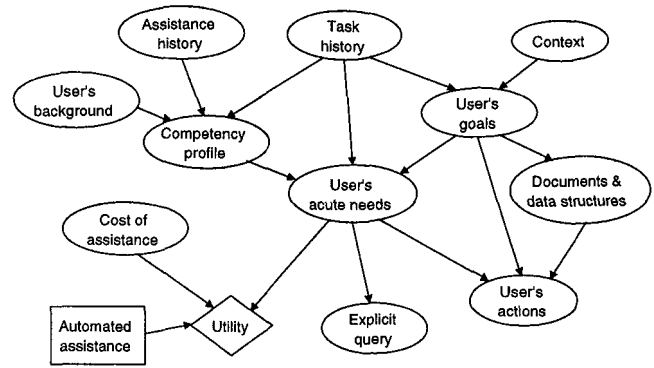

Figure 1: An influence diagram for providing intelligent assistance given uncertainty in a user's background, goals, and competency in working with a software application.

At times, a user may explicitly request assistance. A user's acute needs influence the terms appearing in a user's explicit typed or vocalized queries. As indicated in the influence diagram, the overall goal is to take automated actions to optimize the user's expected utility. A system taking autonomous action to assist users needs to balance the benefits and costs such actions. The value of actions depend on the nature of the action, the cost of the action, and the user's needs. Assessing and integrating such user models would allow a system to compute a probability distribution over a user's informational needs in real time, given a set of observations about activity and explicit queries, when such queries are issued.

## 3 Framing, Constructing, and Assessing Bayesian User Models

Moving from a high-level specification of the problem of Bayesian user modeling to specific domains and prototypes requires a detailed consideration of distinctions and relationships for particular software programs and settings. To better understand the needs and behaviors of users as they encounter problems with the use of a software application, we worked with psychologists at Microsoft's usability laboratory to perform studies with human subjects.

One of the goals of the studies was to gauge the ability of experts to perform the task of guessing users' goals and to provide assistance by simply watching the user's actions through the "keyhole" of the interface. Difficulties noted by experts in the study would be taken as an indication of the difficulty of automating the task of assisting users. We were also interested in identifying evidential distinctions that experts might be using to infer (1) the likelihood that a user needed assistance, and (2) the type of help that was needed given that assistance was desired.

The studies were based on a "Wizard of Oz" design. In the studies, naive subjects with different levels of competence in using the Microsoft Excel spreadsheet application were given a set of spreadsheet tasks. Subjects were informed that an experimental help system would be tracking their activity and would be occasionally making guesses about the best way to help them. Assistance would appear on an adjacent computer monitor. Experts were positioned in a separate room that was isolated acoustically from the users. The experts were given a monitor to view the subjects' screen, including their mouse and typing activity, and could type advice to users. Experts were not informed about the nature of the set of spreadsheet tasks given to the user. Both subjects and experts were told to verbalize their thoughts and the experts, subjects, and the subjects' display were videotaped.

The studies led to several insights. We found that experts had the ability to identify user goals and needs through observation. However, recognizing the goals of users was a challenging task. Experts were typically uncertain about a user's goals and about the value of providing different kinds of assistance. At times, goals would be recognized with an "Aha!" reaction after a period of confusion. We learned that poor advice could be quite costly to users. Even though subjects were primed with a description of the help system as being "experimental," advice appearing on their displays was typically examined carefully and often taken seriously; even when expert advice was off the mark, subjects would often become distracted by the advice and begin to experiment with features described by the wizard. This would give experts false confirmation of successful goal recognition, and would bolster their continuing to give advice pushing the user down a distracting path. Such patterns of poor guesses and "confirmatory" feedback could lead to a focusing on the wrong problem and a loss in efficiency. Experts would become better over time with this and other problems, learning, for example, to becoming conservative with offering advice, using conditional statements (*i.e.*, "if you are trying to do x, then..."), and decomposing advice into a set of small, easily understandable steps.

### 3.1 Identification of Distinctions with Relevance to User Needs

The studies with human subjects helped us to identify several important classes of evidential distinctions. These observational clues appeared to be valuable for making inferences about a user's problems and for making an assessment of the user's need for assistance. The classes of evidence include:

- *Search*: Repetitive, scanning patterns associated with attempts to search for or access an item or functionality. Such distinctions include observation of the user exploring multiple menus, scrolling through text, and mousing over and clicking on multiple non-active regions.

- *Focus of attention*: Selection and/or dwelling on graphical objects, dwelling on portions of a document or on specific subtext after scrolling through the document.

- *Introspection*: A sudden pause after a period of activity or a significant slowing of the rate of interaction.

- *Undesired effects*: Attempts to return to a prior state after an action. These observations include undoing the effect of recent action, including issuing an undo command, closing a dialog box shortly after it is opened without invocating an operation offered in the context of the dialog.

- *Inefficient command sequences*: User performing operations that could be done more simply or efficiently via an alternate sequence of actions or through easily accessible shortcuts.

- *Domain-specific syntactic and semantic content*: Consideration of special distinctions in content or structure of documents and how user interacts with these features. These include domain-specific features associated with the task.

These classes, when subclassed with specific types of data structures and displayed objects, provide a rich set of observations with probabilistic links to a user's goals. In our initial systems, we avoided detailed modeling of deeper commonsense knowledge associated with the syntactic and semantic content of documents and application functionality, and focused initially on the other classes of distinctions.

### 3.2 Structuring Bayesian User Models

Given the results of the user studies, we set out to build and assess Bayesian models with the ability to diagnose a user's needs. We were interested in the quality of inference we might achieve through considering a user's background, ongoing and long-term user actions, data structures, as well as words in a user's query when such a query was issued.

Building effective user models hinges on defining appropriate variables and states of variables. The assessed (or learned) conditional probabilities and the monitoring of users hinge on crisp and appropriate definitions of states of variables. For example, with the use of discrete Bayesian networks, we need to clearly define the specific quantity of time we define to be a "pause after activity."

Figure 2 displays a small Bayesian network that represents the dependency between a pause after activity and the likelihood that a user would welcome assistance. According to the model, a user being in the state of welcoming assistance would shift the probability distribution of observing pauses in activity. The state of desiring assistance also influences the probability of detecting a recent search through multiple menus. We consider also the influence of a user's expertise and the difficulty of a task on the likelihood that a user will need assistance. We also note that a user will pause if he or she is distracted by events unrelated to the user's task. As indicated in Figure 2, we may wish to assert in the Bayesian user model that the difficulty of a task also directly influences the likelihood that a user will become distracted by other events.

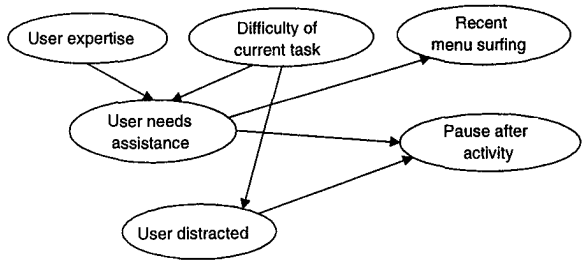

Figure 2: A portion of a Bayesian user model for inferring the likelihood that a user needs assistance, considering profile information as well as observations of recent activity.

We worked with experts to construct, assess, and test Bayesian models for several applications, tasks, and subtasks. Some of our models were built to be "application covering" while others were designed to operate as simpler, context-sensitive agents that would be invoked when specific patterns of activity were observed. As an example, we built and assessed a Bayesian model focused on assisting users with the *Start* button in Windows 95 shell to be invoked when a user was engaging the start button and spending more than a small amount of time navigating within this menu or revisiting the start button multiple times within a predefined horizon. Figure 3 displays a portion of an early application-covering Bayesian network that we constructed for diagnosing a user's goals with the Excel spreadsheet application.

## 4 Temporal Reasoning about User Actions

Bayesian user modeling from a sequence actions over time poses a challenging temporal reasoning problem. Explicit temporal reasoning adds significant complexity to probabilistic representations and inference (Cooper, Horvitz & Heckerman, 1989). Investigation of temporal reasoning with *dynamic Bayesian network* models has focused on problems and approximations for models that consider dependencies among variables within as well as between time slices (Dean & Kanazawa, 1989; Dagum, Galper & Horvitz, 1992; Nicholson & Brady, 1994).

In the general case, we consider temporal dependencies between a user's goals at different times and the user's behavior. Figure 4 captures a Markov representation of the temporal Bayesian user-modeling problem, where we consider dependencies among variables at adjacent time periods. As displayed in the temporal Bayesian network, we include temporal dependencies between goals at the present moment ($Goal_{t_o}$) and at earlier time periods ($Goal_{t_i}$), as well as among observations $E_i$ made at different time periods. Some variables, such as the node labeled *Profile* in Figure 4, capturing the expertise of a user, may change slowly or simply persist over time (as indicated by the arc represented by a broken line).

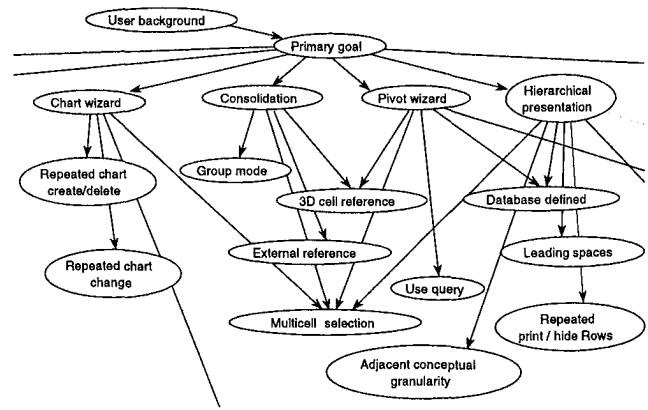

Figure 3: Partial structure of an early formulation of a Bayesian user model for inferring a user's needs for the Excel spreadsheet application.

We studied several approaches to temporal reasoning for user assistance, including dynamic network models and single-stage formulations of the temporal reasoning problem. In the single-stage temporal analyses, we define the target diagnostic problem as inferring the likelihood of alternative goals at the present moment and embed considerations of time within the definitions of observational variables or introduce time-dependent conditional probabilities.

In our earliest Bayesian networks for Lumière/Excel, including the model displayed in Figure 3, we represented and assessed temporal relationships as part of definitions of observations, such as the event, "user performed action $x$ less than $y$ seconds ago." We also experimented with approximations based on direct assessment of parameters of functions that specify the probabilities of observations conditioned on goals as a function of the amount of time that has transpired between the observation and the present moment. The intuition behind the approach is that observations seen at increasingly earlier times in the past have decreasing relevance to the current goals of the user.

The time-dependent probability approach can be viewed as a temporal model-construction methodology. As highlighted in Figure 5 we formulate the problem as a set of single stage inference problems about a user's present goals, and define observations in terms of the amount of time $t$ that has transpired between the present moment and the most recent occurrence of an action, $E_{i,t}$. We assess the conditional probabilities of time-indexed actions, $p(E_{i,t}|Goal_{t_o})$. We found through assessment that a typical temporal dynamic for the conditional probability of observing actions given goals is where the probability decays to the value for the case where the evidence is not observed. That is, the conditional probabilities of observations given goals range from the update associated with an action occurring immediately, $p(E_{i,t_o}|Goal_{t_o})$, to the probability assigned for the case where the observation was not seen within the horizon of the analysis, $p(E_i = false|Goal_{t_o})$. We make the assumption that the rates of decay of likelihoods are independent of one another, conditioned on goals. If an observational variable is influenced by multiple variables, we must assess the decays as a function of all of the states of its parents. For cases where the influence of parent nodes are identified to be causally independent (noisy-or), conditional probabilities can be constructed dynamically from the time-dependent probabilities assessed separately for each parent.

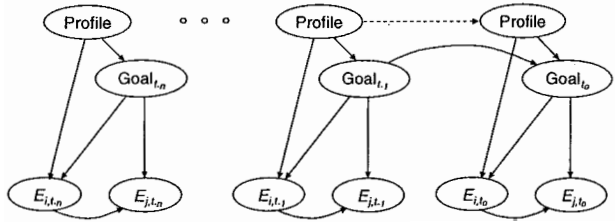

Figure 4: Markov model for temporal reasoning assuming dependencies among the goals of a user in adjacent time periods. A persistent *Profile* variable influences goals and observations in all periods.

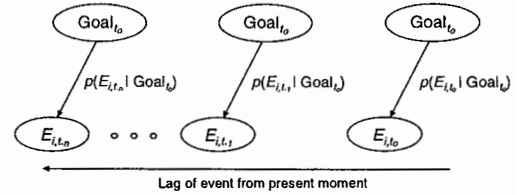

Figure 5: Formulation of the temporal reasoning problem as a set of single-stage problems. We directly assess conditional probabilities of actions as a function of the time that has passed since actions occurred.

In practice, we found it useful to assess the decay of conditional probabilities of actions given current goals with parametric functions. For specially tagged observational variables in the Lumière/Excel model, experts assessed evidential horizon and decay parameters. The evidential horizon is the number of actions or the amount of time that probabilistic relationships persist without change. The decay variables specify the functional class (linear, exponential, etc.) and the parameter defining the details of how the conditional probabilities change after a horizon is reached, as a function of either the amount of time that has passed or the number of actions that have occurred since the observation was noted to become true. For the exponential case, we found it useful to assess a "likelihood half-life" as a function of the number of user actions that had transpired since the event was noted.

## 5 Bridging the Gulf Between System Events and User Actions

To embed Bayesian user models into software applications, it is critical to gain access to a stream of user actions. Unfortunately, systems and applications have not been written with an eye to user modeling. Thus, a critical problem in developing probabilistic and decision-theoretic enhancements for user interface applications is establishing a link between user actions and system events. We found it challenging to gain access to appropriate streams of user and system events. Establishing a rapport with the Excel development team was crucial for designing special instrumented versions of Excel with a usable set of events. Even so, we were limited in the nature of events we could perceive and had to adapt the definition of evidential variables to mesh with available evidence.

A special version of Excel was created that yielded information about subsets of mouse and keyboard actions, as well as information about the status of data structures in Excel files. The events included access to menus being visited and dialog boxes being opened and closed. In addition, we could gain information on the selection of specific objects, including drawing objects, charts, cells, rows, and columns.

We built an events system to establish a fluid link between low-level, *atomic* events and the higher-level semantics of user action we employed in user models. The Lumière events architecture continues to monitor the stream of time-stamped atomic events and to convert these events into higher-level predicates, or *modeled* events representing user actions. We found that transforming system events into such modeled events as "menu surfing," "mouse meandering," and "menu jitter" to be a challenging endeavor. Defining these and other events required detailed analysis of the atomic event streams and iterative composition of temporal functions that could map the atomic event sequences into higher-level observations.

To make the definition of modeled events from atomic events more efficient and flexible, we developed the Lumière Events Language. This temporal pattern recognition language allows atomic events to be used as modeled events directly, as well as for streams of atomic events to be formed into Boolean and set-theoretic combinations of the low-level events. The language also allows system designers to compose new modeled events from previously defined modeled events and atomic events. The events language allowed us to build and modify transformation functions that would be compiled into run-time filters for modeled events. As an example, the following primitives are provided in the event language:

- $Rate(x_i, t)$: The number of times an atomic event $x_i$ occurs in $t$ seconds or commands.

- $Oneof(\{x_1, \ldots, x_n\}, t)$: At least one event of a denoted set of events occurs in $t$.

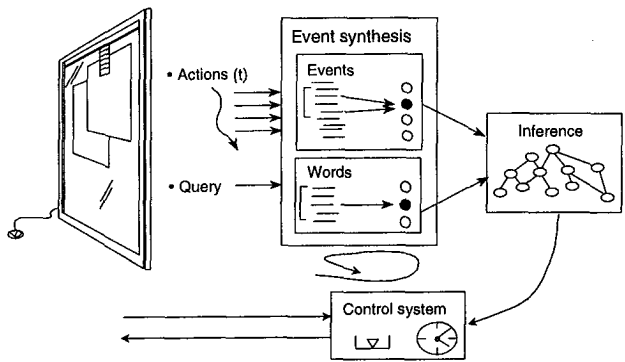

Figure 6: A high-level view of the architecture of Lumière/Excel. Events are transformed into observations represented in the Bayesian model. A control system works to periodically analyze an event queue and perform inference on findings.

- $All(\{x_1,\ldots,x_n\},t)$: All events of a denoted set of events occur at least once in any sequence within $t$.

- $Seq(x_1,\ldots,x_n,t)$: Events occur in a specified order within $t$.

- $TightSeq(x_1,\ldots,x_n,t)$: Events occur in a specified order within $t$ and no other events occur.

- $Dwell(t)$: There is no user action for at least $t$ seconds.

We can use these and other operators in the Lumière Events Language to define filters for higher-level events, such as *user dwelled for at least t seconds following a scroll*. We found it useful to include in the language an efficient means for abstracting sets of low-level events into event classes specified as disjunctions of events (*e.g.*, we specify that a user saving a file via a toolbar icon or keyboard action is to be generalized to *file saved*.) To adapt events defined in terms of durations to different rates of working, we added the ability to expand or contract time to a user-specific *scaled time*, based on detecting the average rate at which commands were executed by users.

## 6  The Lumière/Excel System

The Lumière/Excel prototype for the Excel spreadsheet application was constructed to demonstrate the potential of Bayesian user modeling to the Microsoft Office division. We constructed a Bayesian user model that reasoned about approximately 40 problems spanning a large portion of the Excel application. The Lumière Events Language was used to specify observations as a function of the events emitted by the Excel application. The system was wedded with a Bayesian term-spotting methodology for reasoning about free-text queries (Heckerman & Horvitz, 1998). The Bayesian information retrieval system recognizes approximately 600 terms and considers the probabilistic relevance of these terms to the areas of assistance considered by Lumière/Excel. When queries are available, inference about a user goals, conditioned on a user's actions and the status of data structures, are combined with inference about a user's goals from words. Temporal reasoning was performed with the dynamic temporal modeling approach described in Section 4.

### 6.1  Overall Lumière/Excel Architecture

The overall Lumière/Excel architecture is displayed in Figure 9. Events from the interface are transformed into time-stamped observations. The observations are input to a Bayesian model and a probability distribution over user needs is inferred. If a query is made available, the posterior probabilities from the events system and the Bayesian term-spotting approach are combined through a weighted multiplication to yield a final posterior distribution over needs. Beyond reasoning about the probability distribution over user problems, the system also infers the likelihood that a user needs assistance at the present moment. This probability is used to control the autonomous display of assistance when the system is running behind the scenes. A user-specified probability threshold is used to control the autonomous assistance.

### 6.2  Lumière/Excel Control Policies

We experimented with several overall control schemes for the system. In a *pulsed* strategy the system clock makes a call at a regular interval to a cycle of event analysis, inference, and interface action (depending on the inferred results and the user-specified threshold). In an event-driven control policy, specific sets of atomic and or modeled events are labeled as *trigger* events. When such events occur, a broader events analysis and an inference cycle is triggered. We also experimented with an augmented pulsed approach, where the system calls an analysis cycle at specific time intervals, but also when specific events are noted. In a *deferred analysis*, we attempt to do an analysis at prespecified time intervals but only when idle time is detected. Other opportunities for control include the use of a relatively simple probabilistic or decision-theoretic analysis to make decisions about invoking a larger, less tractable analysis.

### 6.3  Capturing and Harnessing a User Profile

We explored several approaches to custom-tailoring the performance of Lumière/Excel to users with different expertise. In a basic approach to custom-tailoring assistance, we assessed from experts probability distributions over a user's needs for different classes of user expertise, and allowed users to specify their background when using the system.

We also developed the means for updating these probabilities dynamically based on observation of sets of competency *indicator tasks* being completed successfully or specific help topics being accessed and dwelled upon. Beyond analyzing real-time events, Lumière/Excel maintains a persistent *competency profile* in the operating system's registry. The system was given the ability to capture at store at run time sets of key tasks completed and help topics reviewed. This profile can be used to update the probability distribution over needs. In a more general approach, a version of Lumiere was created that enables experts to author special competency variables in the Bayesian user model. These variables have values that reflect the nature and number of times the user demonstrates successful manipulation of different functionalities. These variables can be integrated into the model as any other variable but their states are stored in the persistent user profile.

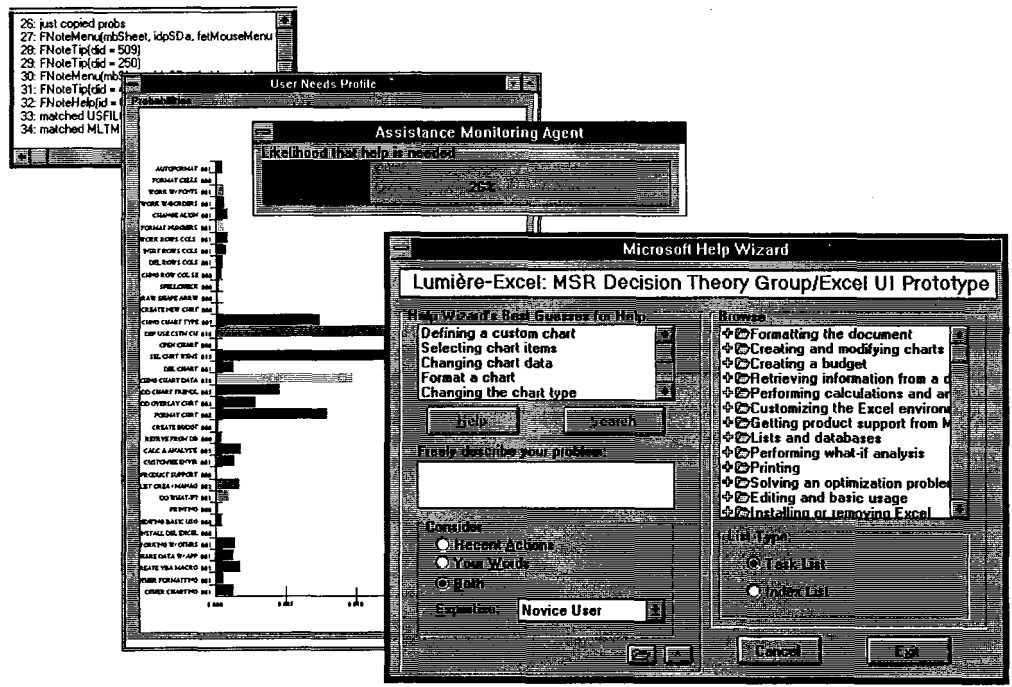

Figure 7: Inference behind the scenes. Components pictured include (from left to right), a display of the atomic event stream, probability distribution over needs, probability that a user would appreciate help at the current moment, and the user interface for the prototype.

### 6.4 Lumière/Excel in Operation

In operation, the Lumière/Excel system continues to monitor events and to update a probability distribution over a user's needs. Given a stream of user events, the system infers needs as well as the overall probability that the user would like assistance immediately. Figure 7 displays a snapshot of Lumière's instrumentation. The small window in the background of the figure displays the stream of atomic events and observations derived from the events. The bar graph in front of the event monitor displays the inferred probability distribution over the needs of the user, given the stream of evidence. Overlayed at the top of the inferred needs, is a bar graph representing the likelihood that the user would like to be notified with some assistance immediately. A user interface for displaying results and interacting with Lumière/Excel appears in the foreground. The left text box of the interface displays recommended assistance sorted by likelihood. A input field allows users to input free-text queries to the system. A menu below the query field allows users to specify their level of competency.

Figure 8 displays the Lumière prototype's user interface, which displays a sorted list of user goals ranked by their probabilities. The figure also highlights the integration of a Bayesian analysis of words in a user's query. Figure 8(a) displays a probability distribution over needs before a query is processed. This probability distribution is computed solely from the user's actions. Figure 8(b) displays an updated probability distribution over needs. This distribution was created by combining the Bayesian action analysis with the output of a Bayesian analysis of terms in the user's free-text query.

If the main assistance interface is not displayed by the user, the system will continue to infer the likelihood a user would like to be offered assistance. When the probability that a user needs assistance exceeds a user-specified threshold, a small window containing the inferred assistance is displayed.

Figure 9 highlights Lumière's autonomous assistance mode. In the example, help was offered autonomously after a user searched through multiple menus, selected the entire spreadsheet, and paused. The assistance window contains a "volume control" which allows users to modify the probability threshold that determines when the system will provide assistance. If the user does not hover over or interact with the autonomous assistance, the window will timeout and disappear after a brief apology for the potential distraction. Recommendations for assistance are noted and the autonomous help will not be offered again until there is a change in the most likely topics. Beyond a user controlled probability threshold, we have experimented with the use of a cost–benefit analysis to control the thresholds at which autonomous assistance is provided, based on the preferences of the user for being alerted.

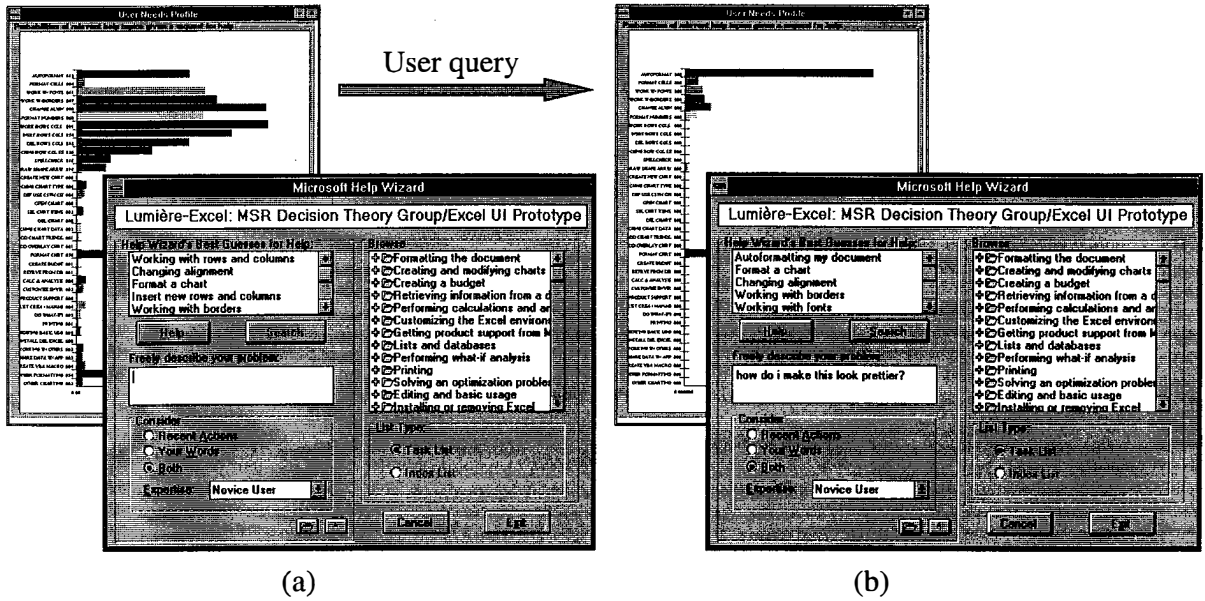

Figure 8: Folding in an analysis of a free-text query into ongoing inference about user needs. (a) Inference results based on actions and (b) revised distribution considering additional information about terms in a user's query.

### 6.5 Beyond Real-Time Assistance

Information about the usage of software can be learned in a variety of settings. The context for most of Lumière research has been in the domain of real-time assistance. However, there is opportunity for using Bayesian user models to design and tailor assistance for offline review. Beyond providing assistance in real time, Lumière was endowed with the ability to track in real-time *patterns* of weakness in skills for using a software application. During a session, the system continues to monitor the likelihoods of user problems. Lumière/Excel tracks the frequency that different problems are encountered during a session and makes recommendations at the end of a session about information the user may wish to review offline. We considered several approaches to identifying the content for a custom-tailored offline tutorial. The method employed in the Lumière/Excel prototype considers, for each area of assistance that is not reviewed during a session, the number of times that the probabilities of user needs for that area exceed a predefined probability threshold. Figure 10 demonstrates the automated generation of recommendations for further reading by Lumière's *help backgrounder*.

## 7 Components of Lumière in the Real World: *Office Assistant*

Our research team has worked closely with the Microsoft Office division at implementing methods developed in Lumière research. The first phase of porting Lumière to the real world occurred with the completion of the Office '97 product suite, containing the *Office Assistant*. Figure 11 displays the interface for the *Office Assistant*. As demonstrated in Figure 11, the Office team committed to a character-based front end to relay the results of Bayesian inference. Compared with Lumière/Excel, the *Office Assistant* employs broader but shallower models reasoning up to thousands of user goals in each Office application. The system uses a rich set of context variables that capture information about the current view and document. However, the system does not employ persistent user profile information and does not reason about competency. Also, the system does not use rich combinations of events over time. Rather, the system only considers a small set of relatively atomic user actions. Furthermore the system employs a small event queue and considers only the most recent events. The system also separates the analysis of words and of events. When words are available, the system does not exploit information about context and recent actions. Finally, the automated facility of providing assistance based on the likelihood that a user may need assistance or on the expected utility of such autonomous action was not employed. Rather, the results of inference are available only when the user requests assistance explicitly. We are continuing to work closely with the Office division and other product groups on the technology transfer of more sophisticated implementations of Bayesian and decision-theoretic user modeling.

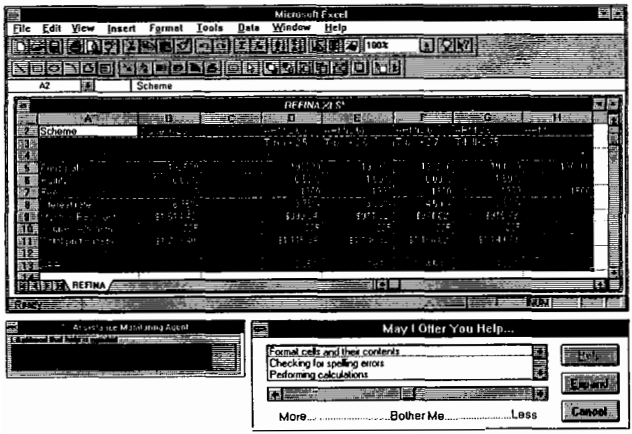

Figure 9: Autonomous display of assistance in Lumière/Excel. Shortly after a user searched through several menus, selected the entire spreadsheet, and paused, the system reaches a probability threshold and posts inferred assistance.

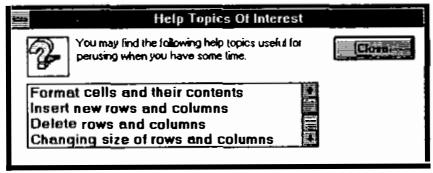

Figure 10: Inference about long-term needs. Lumière/Excel observes patterns of needs in the background and recommends topics for offline perusal at the end of a session.

## 8 Ongoing Research on Bayesian User Modeling

The Lumière/Excel prototype was useful for demonstrating and communicating key ideas on Bayesian user modeling. However, the prototype includes a subset of functionalities that we have been exploring as part of our research on user modeling. Key areas of ongoing work on user modeling include harnessing methods for learning Bayesian models from user log data, integrating new sources of events, and employing automated dialog for engaging users in conversations about goals and needs.

New sources of events can enhance the abilities of a user modeling system to recognize goals and provide appropriate assistance. We have been pursuing opportunities for integrating vision and gaze-tracking into user modeling systems. Even coarse gaze information supplement mouse and keyboard actions with ongoing streams of activity about the attention of a user to regions of a display.

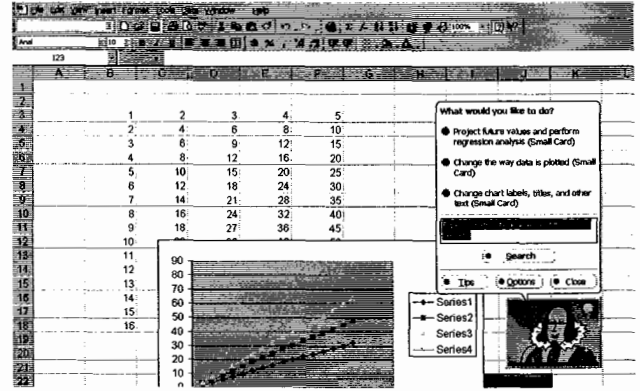

Figure 11: The *Office Assistant*. The fielded *Office Assistant* in the Office '97 suite of applications is based on Lumière research.

We have also been exploring the use of value-of-information computations to engage the user in dialog and to access costly information about user activity and program state. In the early days of Lumière research, we experimented with the use of approximations of value of information to consider the costs and benefits of evaluating previously unobserved variables. At any time in a session, value-of-information identifies previously unobserved variables that would be most valuable to evaluate. In particular, rather than relying on inference to generate a probability distribution over user goals, we can simply ask users about their goals and needs. Our work on Lumière/Excel focused on delivering applications that would simply watch and "listen" for queries rather than making inquiries. However, such dialog can be appropriate and useful.

## 9 Summary and Conclusions

We described the Lumière project with a focus on the Lumière/Excel prototype. We discussed our studies with human subjects to elucidate sets of distinctions that are useful for making inferences about a user's goals and needs and our construction of Bayesian user models. We touched on issues, approximations, and assessment methods for the problem of making inferences from a stream of user actions over time. We presented a basic events-definition language and described an architecture for detecting and making use of events. We also presented our work to integrate evidence from actions and words in a user's query. We reviewed work on autonomous decision making about user assistance controlled by a user-specified probability threshold. Finally, we discussed real-time inference to support the custom-tailoring of tutorial materials for review in offline settings.

Lumière/Excel and its descendant, the *Office Assistant*, represent initial steps in making software applications more intuitive about the goals and needs of users. The dominance of uncertainty in understanding and supporting the goals of people highlights the rich opportunities ahead for harnessing probability and utility at the computer–human interface.

## Acknowledgments

We are indebted to Sam Hobson for his enthusiastic support of Bayesian user modeling in Office applications, and to Kristin Dukay and Eric Hawley for leading teams of usability experts down unexplored territory (against Kristin's better senses). Leah Kaufman managed the Wizard of Oz studies and later studies with users of the Office Assistant. We also acknowledge the indefatigable efforts of Adrian Klein and Eric Finkelstein who joined Sam in transforming components of Lumière into the Office Assistant and spurring the creation of an intelligent-assistance culture in the Office product group. Chris Meek reviewed an earlier draft of this paper.

## References

Albrecht, D., Zukerman, I., Nicholson, A., and Bud, A. (1997). Towards a Bayesian model for keyhole plan recognition in large domains. In *Proceedings of the Sixth International Conference on UserModeling, Sardinia, Italy*, pages 365–376. User Modeling, Inc., Springer-Verlag.

Charniak, E. and Goldman, R. (1993). A Bayesian model of plan recognition. *Artificial Intelligence*, 64(1):53–79.

Conati, C., Gertner, A., VanLehn, K., and Druzdzel, M. (1997). Online student modeling for coached problem solving using Bayesian networks. In *Proceedings of the Sixth International Conference on User Modeling, Sardinia, Italy*, pages 231–242. User Modeling, Springer-Verlag.

Cooper, G. and E. Horvitz, D. Heckerman, R. C. (1988). Conceptual design of goal understanding systems: Investigation of temporal reasoning under uncertainty. Technical Report Technical Memorandum NAS2-12381, Search Technology and NASA-Ames Research Center, Mountain View, CA.

Cooper, G., Horvitz, E., and Heckerman, D. (1989). A method for temporal probabilistic reasoning. Technical Report KSL-88-30, Knowledge Systems Laboratory, Stanford University, Stanford, CA.

Dagum, P., Galper, A., and Horvitz, E. (1992). Dynamic network models for forecasting. In *Proceedings of the Eighth Workshop on Uncertainty in Artificial Intelligence*, pages 41–48, Stanford, CA. Association for Uncertainty in Artificial Intelligence.

Dean, T. and Kanazawa, K. (1989). A model for reasoning about persistence and causation. *Computational Intelligence*, 5(3):142–150.

Heckerman, D., Breese, J., and Rommelse, K. (1995). Decision-theoretic troubleshooting. *CACM*, 38:3:49–57.

Heckerman, D. and Horvitz, E. (1998). Inferring informational goals from free-text queries: A bayesian approach. In *Proceedings of the Fourteenth Conference on Uncertainty in Artificial Intelligence*. AUAI, Morgan Kaufmann.

Heckerman, D., Horvitz, E., and Nathwani, B. (1992). Toward normative expert systems: Part I. The Pathfinder project. *Methods of information in medicine*, 31:90–105.

Horvitz, E. (1997). Agents with beliefs: Reflections on Bayesian methods for user modeling. In *Proceedings of the Sixth International Conference on UserModeling, Sardinia, Italy*, pages 441–442. User Modeling, Springer-Verlag.

Horvitz, E. and Barry, M. (1995). Display of information for time-critical decision making. In *Proceedings of the Eleventh Conference on Uncertainty in Artificial Intelligence*, pages 296–305, Montreal, Canada. Morgan Kaufmann, San Francisco, CA.

Horvitz, E., Breese, J., and Henrion, M. (1988). Decision theory in expert systems and artificial intelligence. *International Journal of Approximate Reasoning*, Special Issue on Uncertain Reasoning, 2:247–302.

Horvitz, E., Heckerman, D., Nathwani, B., and Fagan, L. (1984). Diagnostic strategies in the hypothesis-directed Pathfinder system. In *Proceedings of the First Conference on Artificial Intelligence Applications, Denver, CO*, pages 630–636. IEEE.

Jameson, A. (1996). Numerical uncertainty management in user and student modeling: An overview of systems and issues. *User Modeling and User-Adapted Interaction*, 5:193–251.

Nicholson, A. and Brady, J. (1994). Dynamic belief networks for discrete monitoring. *IEEE Transactions on Systems, Man, and Cybernetics*, 24(11):1593–1610.

Pynadath, D. and Wellman, M. (1995). Accounting for contenxt in plan recognition with application to traffic monitoring. In *Proceedings of the Eleventh Conference on Uncertainty in Artificial Intelligence*, pages 472–481, Montreal, Canada. Morgan Kaufmann, San Francisco, CA.